\documentclass{article}

\PassOptionsToPackage{numbers}{natbib}
\usepackage{graphicx}

    \usepackage[preprint]{neurips_2021}

\usepackage[utf8]{inputenc} 
\usepackage[T1]{fontenc}    
\usepackage{hyperref}       
\usepackage{url}            
\usepackage{booktabs}       
\usepackage{amsfonts}       
\usepackage{nicefrac}       
\usepackage{microtype}      
\usepackage{xcolor}         
\usepackage{duckuments}
\usepackage{amsmath}

\title{On the Bias Against Inductive Biases}

\author{%
  George Cazenavette \\
  Robotics Institute\\
  Carnegie-Mellon University\\
  Pittsburgh, PA 15213 \\
  \texttt{gcazenav@cs.cmu.edu} \\
   \And
  Simon Lucey\\
  Robotics Institute\\
  Carnegie-Mellon University\\
  Pittsburgh, PA 15213 \\
  \texttt{slucey@cs.cmu.edu} \\
}

\begin{document}

\bibliographystyle{plainnat}

\maketitle

\begin{figure}[h]
    \centering
    \includegraphics[width=\linewidth]{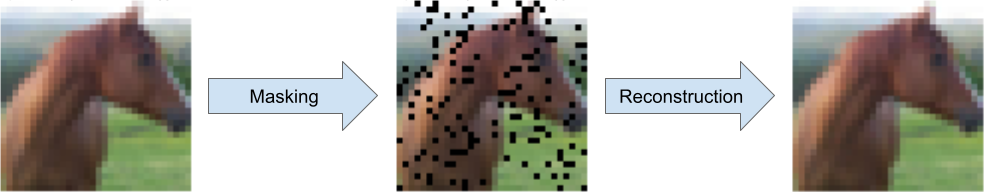}\\
    \includegraphics[width=\linewidth]{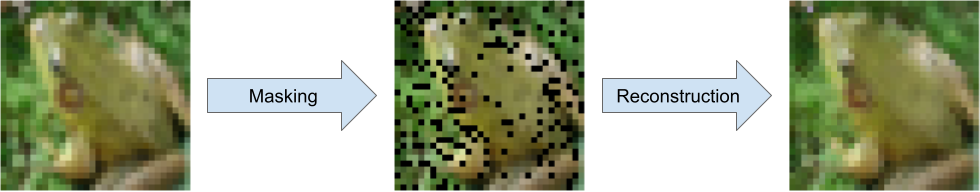}\\
    \includegraphics[width=\linewidth]{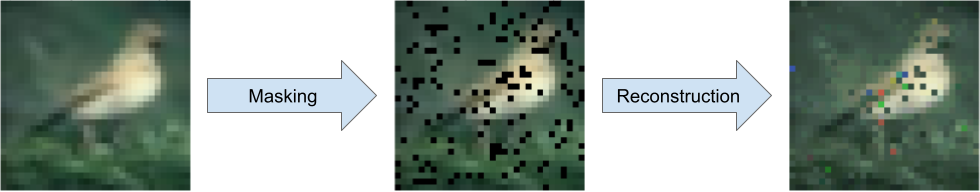}
    \caption{BERT reconstructions using relatively shallow \textbf{convolutional} (top), \textbf{MLP-mixer} (middle), and \textbf{transformer} (bottom) networks. Shallow transformer networks consistently suffer in their reconstructions, hinting towards the benefits of inductive biases in moderately-sized networks.}
    \label{fig:cover}
\end{figure}
\begin{abstract}
   Borrowing from the transformer models that revolutionized the field of natural language processing, self-supervised feature learning for visual tasks has also seen state-of-the-art success using these extremely deep, isotropic networks. However, the typical AI researcher does not have the resources to evaluate, let alone train, a model with several billion parameters and quadratic self-attention activations. To facilitate further research, it is necessary to understand the features of these huge transformer models that can be adequately studied by the typical researcher. One interesting characteristic of these transformer models is that they remove most of the inductive biases present in classical convolutional networks. In this work, we analyze the effect of these and more inductive biases on small to moderately-sized isotropic networks used for unsupervised visual feature learning and show that their removal is not always ideal.
\end{abstract}


\section{Introduction}

With the release of the seminal paper ``Attention is all You Need'' \cite{attention}, the field of natural language processing transformed into an area completely dominated by transformer architectures capable of distilling the world's massive amount of unlabeled data into a formidable knowledge base. Gone were the days of recurrent and even convolutional neural networks as the attention-based transformer model lead to such advancements as BERT \cite{bert} and the GPT series of models \cite{gpt, gpt2, gpt3}. The overwhelming success of these massive isotropic (maintaining the same representation shape through all layers) models left researchers wondering if the same techniques could be applied to computer vision, leading to the development of the image-GPT model \cite{igpt}. 

In addition to the isotropic architecture and unsupervised training objective, the image-GPT model is also unique among vision models in that it eliminates most, if not all, of the inductive biases traditionally associated with computer vision problems. Instead of continuous pixel values as input, image-GPT discretizes 24-bit RGB pixel values by clustering them into 9-bit one-hot vectors. Rather than locality-considering convolutions, image-GPT performs position-agnostic self-attention operations. 

Instead of relying on these inductive biases, image-GPT's extreme parameter count and depth allow it to learn these critical relationships that come inherent with the natural world from scratch. With the \textit{smallest} model proposed in the paper \cite{igpt} having 76 million parameters and the \textit{largest} having 6.8 billion coupled with the extra-quadratic memory usage of the self-attention operation, there is unfortunately no way for the typical AI researcher to evaluate, let alone train these massive models.

To facilitate the advancement of AI research and avert a potential AI winter, it is necessary to distill advancements made capable by massive amounts of computing power into their core ideas that can be further studied by the community at-large. As such, it is critical to understand the features of these huge transformer models that can be adequately studied by the typical AI researcher. In this work, we explore small to moderately-sized networks in the style of image-GPT capable of being trained with our modest computing resources and examine the effects of including, removing, or even replacing the inductive biases excised by image-GPT.

\section{Related Works}
Given that the world is full of immeasurable amounts of unlabeled data, the goal of unsupervised representation learning is quite attractive. While most of the recent advancements in unsupervised learning have spawned, computer vision researchers have also developed methods of training on unlabeled data through self-supervised learning.
\paragraph{Unsupervised Natural Language Processing}
While the transformer model was originally proposed for paired sequence to sequence translation \cite{attention}, it was quickly adapted to the BERT objective to learn representations in a fully unsupervised manner by masking out certain tokens (words) in the input sequence with the goal of predicting these same tokens in the output sequence \cite{bert}. The transformer model was also used to learn representations by training on the unsupervised auto-regressive objective wherein the model simply had to predict the next token in an incomplete sequence, leading to the development of the Generative Pre-Training (GPT) series of models \cite{gpt, gpt2, gpt3}. It was these two unsupervised objectives coupled with the transformer architecture that eventually lead to the development of image-GPT: a Generative Pre-Training model for images that is functionally the same to BERT or GPT in that it simply treats pixels as words.

\paragraph{Contrastive Visual Learning}
An aspect of computer vision that does not translate to natural language processing is the idea of data-augmentation or applying arbitrary transformations to an image. The idea of data augmentation lead to one of the most popular methods of self-supervised representation learning for vision: contrastive learning \cite{simclr}. In this method of representation learning, sequences of specific stochastic transformations are applied independently. The model then compares embeddings of two transformed images. If they come from the same source image, then the model minimizes the distance between their embeddings, and if not, the the model maximizes the same distance. Since this method effectively assigns pseudo-labels for a discriminative task, it is considered to be self-supervised learning rather than pure unsupervised learning.

\section{Model Architectures}
\begin{figure}
    \centering
    \begin{tabular}{c c c }
    \multicolumn{3}{c}{Distance Preservation in Input Encodings}\\
    \\
       \includegraphics[width=0.3\linewidth]{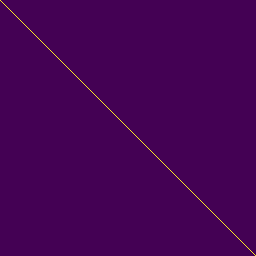} & \includegraphics[width=0.3\linewidth]{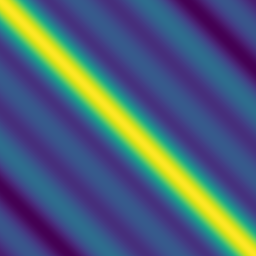} & \includegraphics[width=0.3\linewidth]{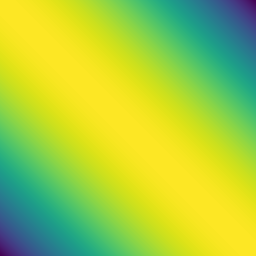}\\
       \\
       One-Hot & Random Fourier ($\sigma=1$) & Continuous
    \end{tabular}
    \caption{Each pixel represents the cosine similarity between the encodings of the two pixel values at at (x,y) position with the top left representing (0,0) and the bottom right (255, 255). A one-hot representation destroys all distance while a continuous representation perfectly preserves it. A random fourier embedding allows us to preserve distance in a local band while also substantially increasing rank.}
    \label{fig:distance_preservation}
\end{figure}

While the image-GPT model exclusively uses transformer blocks in its layers, we also explore two other types of isotropic blocks: mixer and convolutional. We also consider different methods of encoding the input that each carry their own level of inductive bias. Furthermore, image-GPT was trained on both the auto-regressive and BERT \cite{bert} objectives while here, we only focus on the latter.

\subsection{Block Types}

Like image-GPT, all of our models are isotropic in shape such that they keep the same size representation throughout all layers. After an initial layer to encode the input to the desired dimensionality, the models then consist of $L$ blocks. In this work, we explore networks consisting of three types of blocks: transformer blocks \cite{attention}, MLP-mixer blocks \cite{mixer}, and traditional convolutional blocks.

For comparison's sake, the three block types are kept as similar as possible such that they have the same number of non-linearities and the same activation dimensionality.

\paragraph{Convolutional Block}
Each type of block contains a different level of inductive bias. The convolutional block carries the locality bias that comes with the filter in the convolution operation itself. The forward pass of the convolutional block $\mathbf{X} \to \mathbf{Y}$ with intermediate skip-connection \cite{resnet} point $\mathbf{U}$ can be described by the following equations:
\begin{equation}
\begin{split}
    \mathbf{U} &= \mathbf{X} +  \mbox{GELU}(\mbox{Conv2D} (\mbox{LayerNorm}(\mathbf{X}), \mathbf{W}_1))\\
    \mathbf{Y} &= \mathbf{U} + \mbox{GELU}(\mbox{Conv2D} (\mbox{LayerNorm}(\mathbf{U}), \mathbf{W}_2))
\end{split}
\end{equation}
where GELU is the Gaussian Error Linear Unit activation \cite{gelu} as used in image-GPT.

\paragraph{Mixer Block}We adapt this block from the recently proposed MLP-Mixer architecture \cite{mixer}. The mixer block is similar to the transformer block except that it performs a token-mixing MLP instead of the self-attention operation. As such, it still retains the long-range connections that make self-attention desirable without the extra-quadratic activation memory cost. The mixer block carries a positional bias in its token-mixing MLP as the MLP carries different weights for each token position in the representation. The forward pass of the the mixer block $\mathbf{X} \to \mathbf{Y}$ (where $\mathbf{X}, \mathbf{Y}$ are 2D representations made of column-stacked tokens) is described by:
\begin{equation}
\begin{split}
    \mathbf{U} &= \mathbf{X} + \mathbf{W}_2 \mbox{GELU}(\mathbf{W}_1 \mbox{LayerNorm}(\mathbf{X}))\\
    \mathbf{Y} &= (\mathbf{U}^T + \mathbf{W}_4\mbox{GELU}(\mathbf{W}_3\mbox{LayerNorm}(\mathbf{U})^T))^T
\end{split}
\end{equation}
\paragraph{Transformer Block}Lastly, the transformer carries no such bias as the self-attention operator is completely agnostic to the position of its input tokens. The transformer is also typically a gatekeeper for training hardware due to the quadratic memory usage of the self-attention activations. The forward pass of the transformer block $\mathbf{X} \to \mathbf{Y}$ (where $\mathbf{X}, \mathbf{Y}$ are again 2D representations made of column-stacked tokens) is described by:
\begin{equation}
\begin{split}
    \mathbf{U} &= \mathbf{X} +   \mbox{MultiHeadAttention}(\mbox{LayerNorm}(\mathbf{X}), \mathbf{W}_Q, \mathbf{W}_K, \mathbf{W}_V)\\
    \mathbf{Y} &= (\mathbf{U}^T + \mathbf{W}_2\mbox{GELU}(\mathbf{W}_1\mbox{LayerNorm}(\mathbf{U})^T))^T
\end{split}
\end{equation}
\subsection{Input Encodings}
In this work, we experiment with three different input encodings: one-hot pix2vec, continuous space, and random fourier features. The image-GPT model learns a one-hot pix2vec embedding from scratch where any and all preserved distance from the input space must be learned by the model itself. While this idea may seem counter-intuitive at first, giving the network a one-hot input seeds it with a representation that is already of maximal rank, allowing for easier discriminative tasks down-stream. With our other input encodings, we aim to further explore this trade-off between distance preservation and rank escalation.

\paragraph{pix2vec}
Just like the word2vec embeddings used in language models, image-GPT learns a pix2vec embedding from one-hot encoded pixels to a vector space matching the latent dimensionality. To do this, image-GPT clusters all possible 24-bit RGB pixel values into a 9-bit one-hot space. Instead, we learn a separate 8-bit pix2vec embedding for each channel. In this case, the model must learn any required distance preservation separately for each channel.

\begin{figure}[]
    \centering
    \begin{tabular}{c c c}
        \multicolumn{3}{c}{Distance Preservation in Random Fourier Encodings}\\
        \\
       \includegraphics[width=0.3\linewidth]{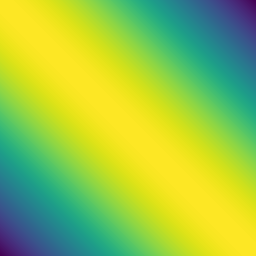} & \includegraphics[width=0.3\linewidth]{fig/rff_1.png} & \includegraphics[width=0.3\linewidth]{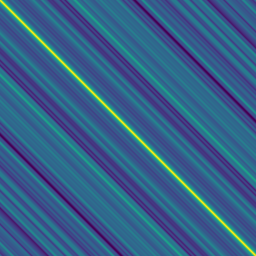}\\
       \\
       $\sigma = 0.1$ & $\sigma = 1$ & $\sigma = 10$
    \end{tabular}
    \caption{Each pixel represents the cosine similarity between the encodings of the two pixel values with the top left representing (0,0) and the bottom right (255, 255). If $\sigma$ is too large, we lose all notion of distance. If too small, we perfectly preserve distance but have no rank expansion. A value of $\sigma=1$ allows us to preserve distance locally while still increasing rank.}
    \label{fig:fourier_spectrum}
\end{figure}

\paragraph{Random Fourier Features}Inspired by the usage of positional encoding via exponential fourier features in NeRF \cite{mildenhall2020nerf}, we experiment by encoding the pixel values themselves with random fourier features. Borrowing notation from the NeRF paper, the $k$-frequency random fourier feature encoding is formally described by the following function:
\begin{equation}
\begin{split}
    \gamma(p) = (\sin(2\beta_0 \pi p), \cos(2\beta_0 \pi p), \hdots, \sin(2\beta_{k-1} \pi p), \cos(2\beta_{k-1} \pi p))
\end{split}
\end{equation}
where $\beta_i \sim \mathcal{N}(0, \sigma^2)$. By using random fourier features instead of exponential fourier features, we can have arbitrarily many frequencies without destroying all distance. Here, $\sigma$ directly acts as a slider between distance preservation and rank escalation as seen in Figure \ref{fig:fourier_spectrum}. 
\paragraph{Raw Continuous Input}
For our continuous input, we do not perform any per-pixel normalization. Instead, all 8-bit pixel values are simply mapped to the range $[-1, 1]$ by subtracting 127.5 and then dividing by the same. This method of encoding perfectly preserves all distance between pixel values but does not provide an inherent rank escalation.

As stated in the original BERT paper \cite{bert}, the acronym stands for \textbf{B}idirectional \textbf{E}ncoder \textbf{R}epresentations from \textbf{T}ransformers. To refer to the masked token prediction as the BERT objective even when discussing non-transformer architectures may be a potential misnomer, but we will continue referring to it as such for simplicity.

While other methods of self-supervised learning in vision rely heavily on pseudo-labels or input transformations, the BERT objective originates from natural language processing where input transformations do not make sense. As such, the BERT objective provides a goal for pure unsupervised learning by simply requiring the prediction of the masked parts of the input. 

To describe the BERT objective, let $M$ be the set of masked input indices where each input index $i$ has a set chance of being in $M$. Keeping with precedence \cite{bert, igpt}, we fix this chance at 15\%. Then the  training objective is to minimize the negative log-likelihood of the predicted discrete values of the masked pixels given the unmasked pixels. Formally, over the training set $X$, the BERT objective is to minimize
\begin{equation}
\begin{split}
    \mathcal{L}_{BERT} = \mathbb{E}_{x \sim X} \mathbb{E}_M \sum_{i\in M}[-\log p(x_i | x_{[1,n]\setminus M})]
\end{split}
\end{equation}
Note that by this objective, our model must predict the discrete, one-hot encoded pixels regardless of the input encoding. While this is a much harder objective than a regression, it prevents the model from simply predicting the average of the surrounding pixels and, in doing so, forces the model to learn a meaningful representation of the input data.
\section{Experiments}
For our experiments, we implement our models and training procedures in TensorFlow \cite{tensorflow2015-whitepaper} using the higher-lever Keras wrapper \cite{chollet2015keras}. Our training hardware consists of 4 NVIDIA TITAN Xp GPUs with 12GB of VRAM each for a total of 48GB of VRAM.
\subsection{Dataset}
In the image-GPT paper, they train their models on the unsupervised objective using the entirety of the \char`\~ 1.3 million sample ImageNet dataset \cite{deng2009imagenet} before evaluating it via linear probing and fine-tuning for ImageNet classification. Given our modest amount of compute and our focus on such scenarios, we instead train our unsupervised model on CIFAR-10 \cite{cifar} and evaluate exclusively with linear probing on the same dataset.

\subsection{Training Procedure}
To pre-train our models on the unsupervised BERT objective, we use the Adam optimizer \cite{kingma2014adam}. We begin with one epoch of linearly warming-up the learning rate from 0 to 0.01 before decaying back to 0 over the next 50 epochs using a cosine schedule \cite{cosine}.

After we have pre-trained on the unsupervised task, we begin training linear probes to evaluate the efficacy of the learned representation. We use the same optimizer and learning rate as when training the unsupervised models, but we instead decay over 100 epochs or until convergence.

Unlike classical auto-encoders that contain an information bottleneck layer, our models are isotropic, so there is not an immediately clear answer as to which layer would provide the best features. The image-GPT paper also notes that their best unsupervised features come from a combination of several layers. To avoid this combinatoric search space of training given our modest amount of compute, we will simply consider the linear probes from each isolated layer. Since we are analyzing the effects of different inductive biases instead of reaching for state-of-the-art results, this seems okay for now, but it would be interesting to explore this avenue in future work.

For all except the 4-headed transformer models, we use a batch size of 128. For the 4-headed transformer models, we had to reduce the batch size to 32 to fit on our GPUs.

\subsection{Selected Models}
For all our models, we choose a latent channel dimension of 128. We experiment with our various input encodings (pix2vec, random fourier features, raw continuous) and block types (convolutional, mixer, transformer) for networks of size (number of blocks) 1, 3, and 6. However, training the multi-headed transformer models and then linearly probing each layer became prohibitively expensive with our hardware once we reached size 6, so we halted this training after the first few to focus on other settings.
\begin{figure}[]
    \centering
    \begin{tabular}{c c}
        \includegraphics[width=0.48\linewidth]{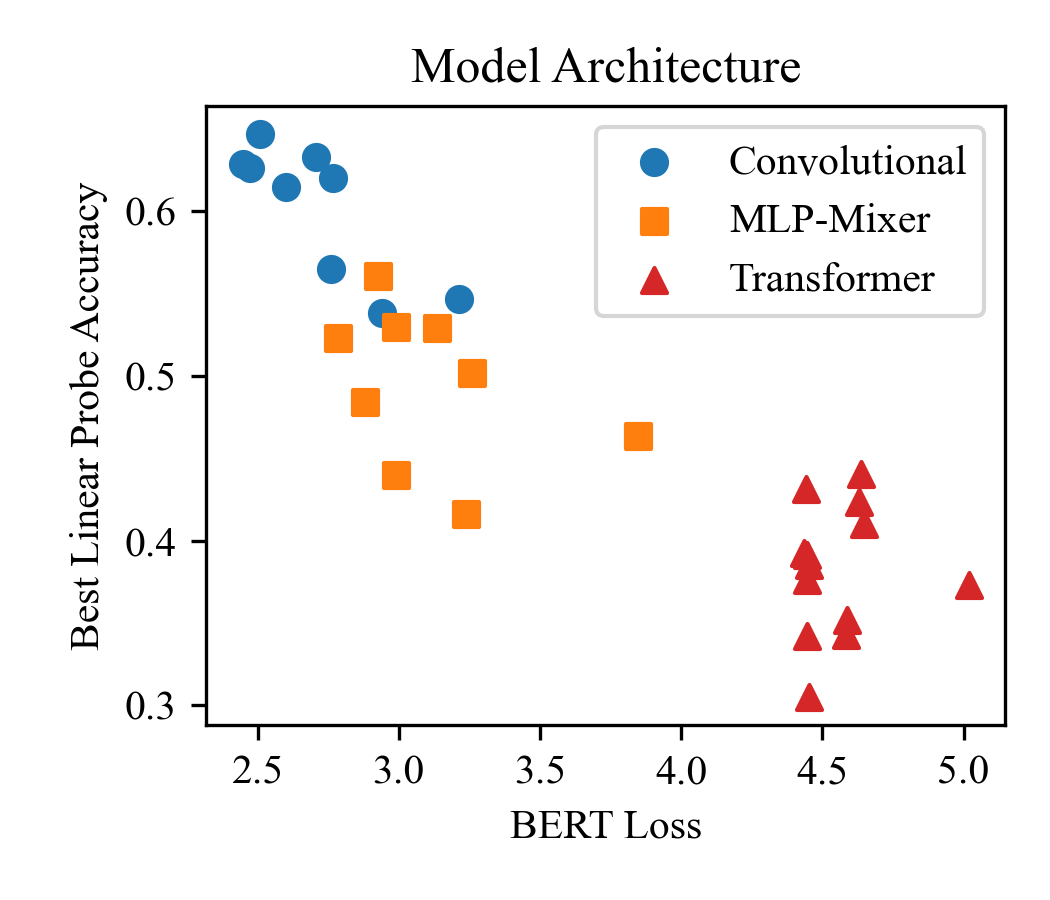} & \includegraphics[width=0.48\linewidth]{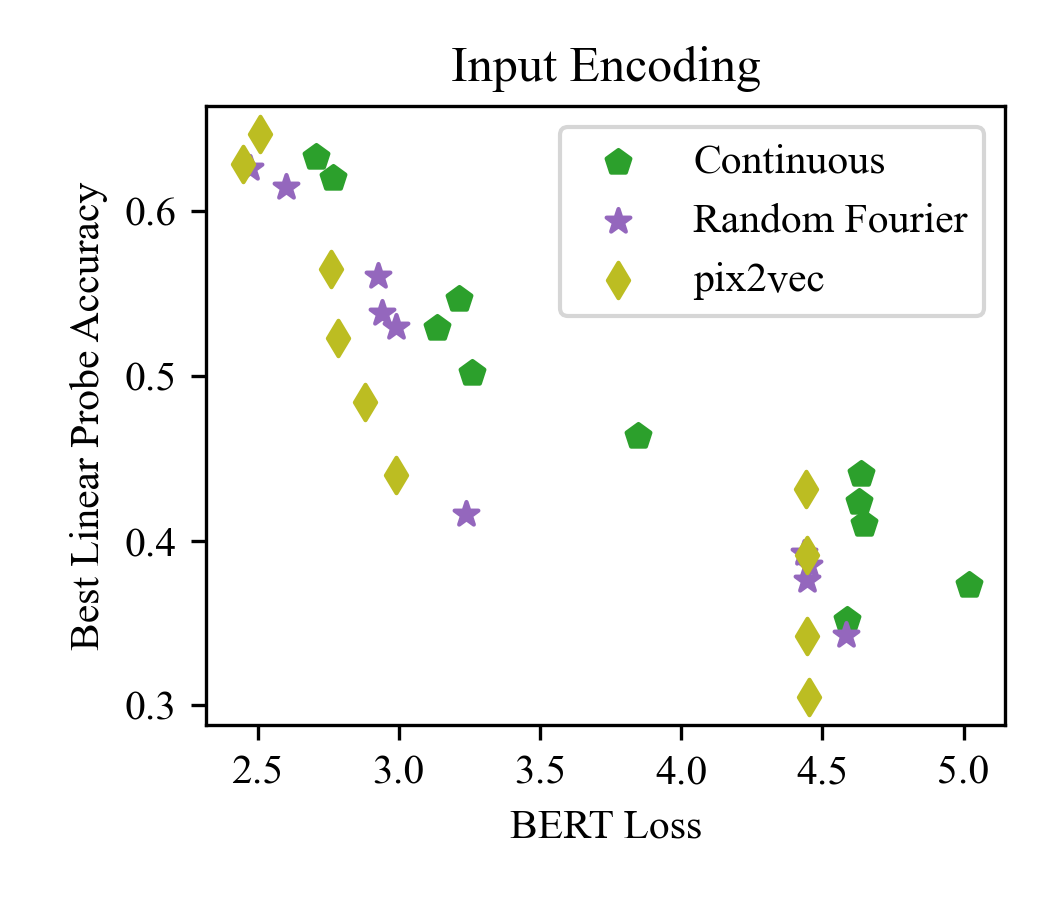}
    \end{tabular}
    \caption{(Left) By labeling the unsupervised loss vs classification accuracy points by model type, we clearly see that convolutional networks perform the best on both the unsupervised and classification tasks followed by MLP-mixer networks and distantly trailed by transformer networks. (Right) By labeling the same points by input encoding type instead, we see a trend where for a fixed unsupervised loss, continuous input yields the best classification accuracy, followed by random fourier features, and finally pix2vec.}
    \label{fig:scatters}
\end{figure}

\section{Results}
We present our results on networks of size 1, 3, and 6. Results for networks of size 12 can be found in the supplementary material.

\subsection{Architecture Class}
\begin{figure}[]
    \centering
\begin{tabular}{c c c c}
    \multicolumn{4}{c}{Distance Preservation in Learned pix2vec Embeddings}\\
    \\
     \includegraphics[width=0.22\linewidth]{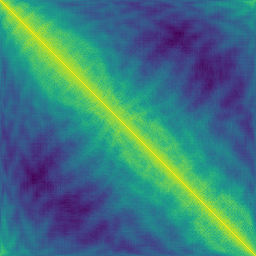} & \includegraphics[width=0.22\linewidth]{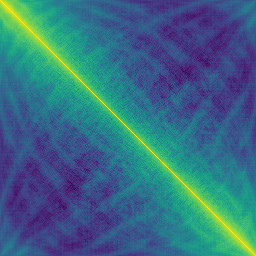} & 
     \includegraphics[width=0.22\linewidth]{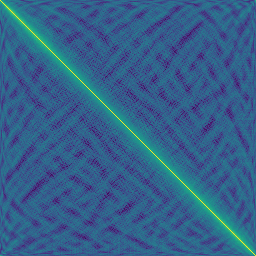} & \includegraphics[width=0.22\linewidth]{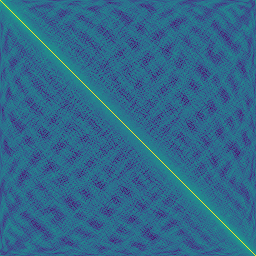}\\
     \\
     Convolution & MLP-Mixer & 1-Head Transformer & 4-Head Transformer
\end{tabular}
    \caption{(size-3 networks, red channel) The different types of model architectures resulted in very different levels of distance preservation in the pix2vec embeddings. Relative performance of said models correlates with the level of distance preservation shown here.}
    \label{fig:pix2vec_embeddings}
\end{figure}
Looking at the left part of Figure \ref{fig:scatters}, we see that transformer networks consistently have the worst performance on the unsupervised task while convolutional networks perform the best. As noted by the image-GPT authors, we see a strong correlation between the unsupervised generative model's performance and the accuracy of the best linear probe. As such, we also see that the transformer networks also perform the worst on the classification task while convolutional networks again perform the best, having by far the best accuracies.

We can see the direct result of the different model classes unsupervised validation loss in Figure \ref{fig:cover}. In the convolutional setting (top), we see that the horse is nearly perfectly reconstructed by the network. In the MLP-mixer setting (middle), we see all the colors of in the reconstructed frog seem to be from the correct palette, but some of them are still misplaced. Lastly, in the transformer setting, we see the model repeatedly predict completely incorrect colors in the reconstructed bird, explaining the transformer models' consistently much higher unsupervised loss.

If we look at only models that used the pix2vec input embeddings, we can gain further insight to this phenomenon. Figure \ref{fig:pix2vec_embeddings} shows the learned pix2vec embeddings for the various types of architectures. We see that the level of distance preservation is highest in the convolutional models with that of the MLP-mixer models being slightly worse. However, the distances learned by the transformer models degrade to almost random patterns aside from a \textit{very} narrow local band just off the main diagonal. Note that it is not possible for there to be 0 similarity between pixel values here as in Figure \ref{fig:distance_preservation} because the learned embedding is an under-complete transformation of the one-hot encoding. In other words, it is not possible to have a set of 256 orthogonal vectors in a 128 dimensional space.

While there is certainly more going on in later layers in the network, it is interesting to note that the level of distance preservation in the learned pix2vec embedding strongly correlates with the relative performance of those model classes as a whole.

\subsection{Encoding Method}
\begin{figure}[]
    \centering
    \begin{tabular}{c c}
        \includegraphics[width=0.48\linewidth]{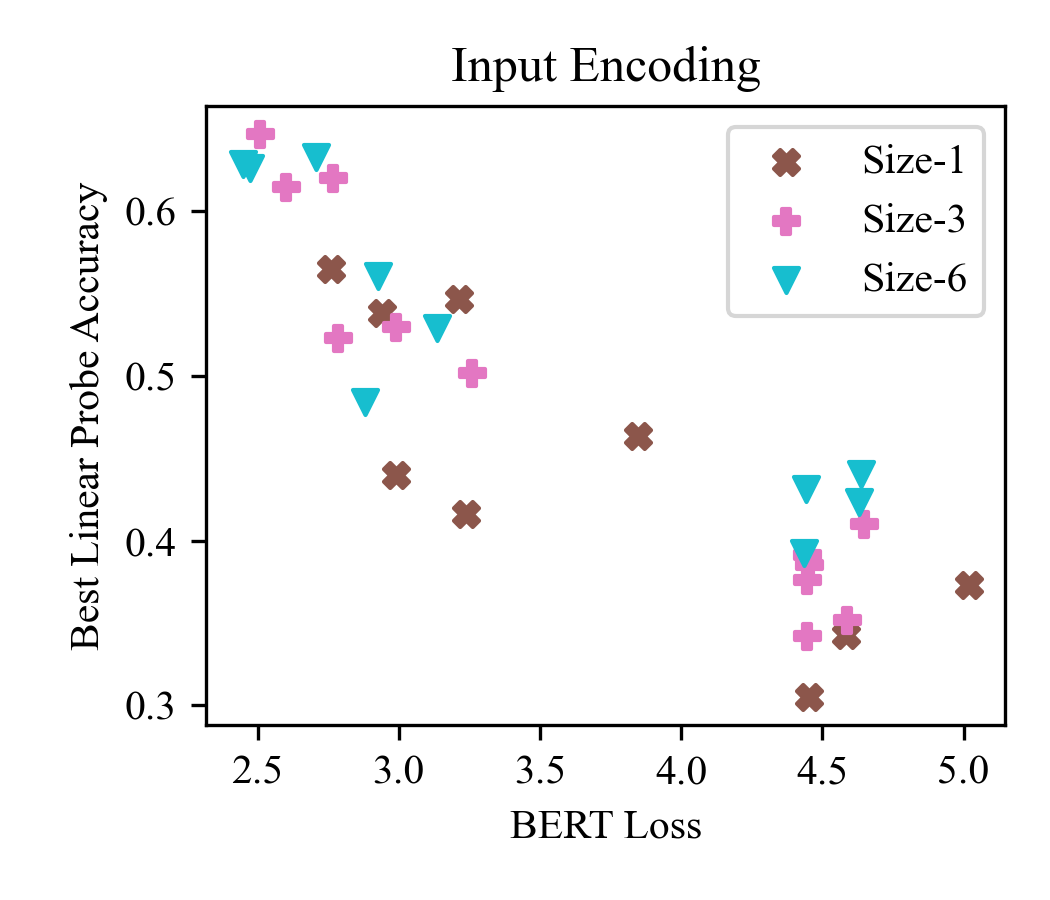} & \includegraphics[width=0.48\linewidth]{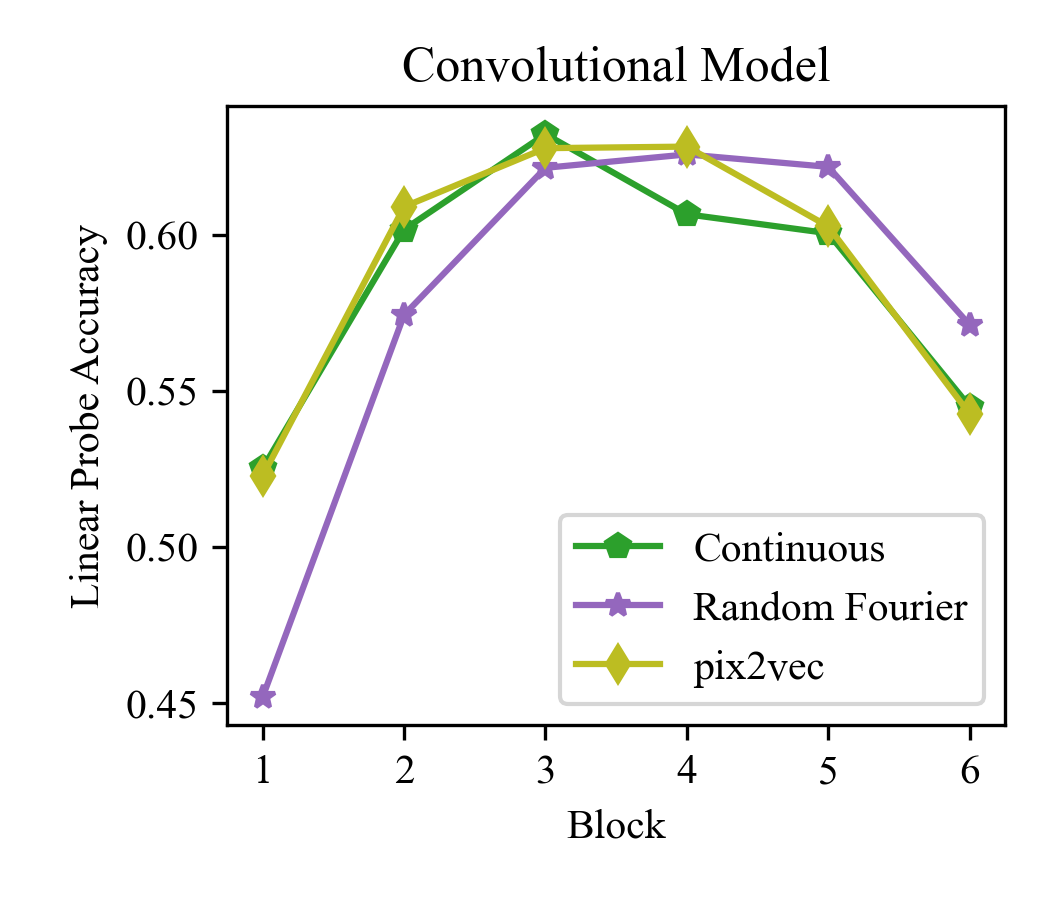}
    \end{tabular}
    \caption{(Left) By labeling the unsupervised loss vs classification accuracy points by model size, we see (unsurprisingly) that deeper models tend to have better performance than shallow ones. (Right) We see that for convolutional models, the best single linear probe comes from the middle of the network, just as observed in the transformer models of image-GPT \cite{igpt}.}
    \label{fig:scatters}
\end{figure}
\begin{figure}[]
    \centering
    \begin{tabular}{c c}
        \includegraphics[width=0.48\linewidth]{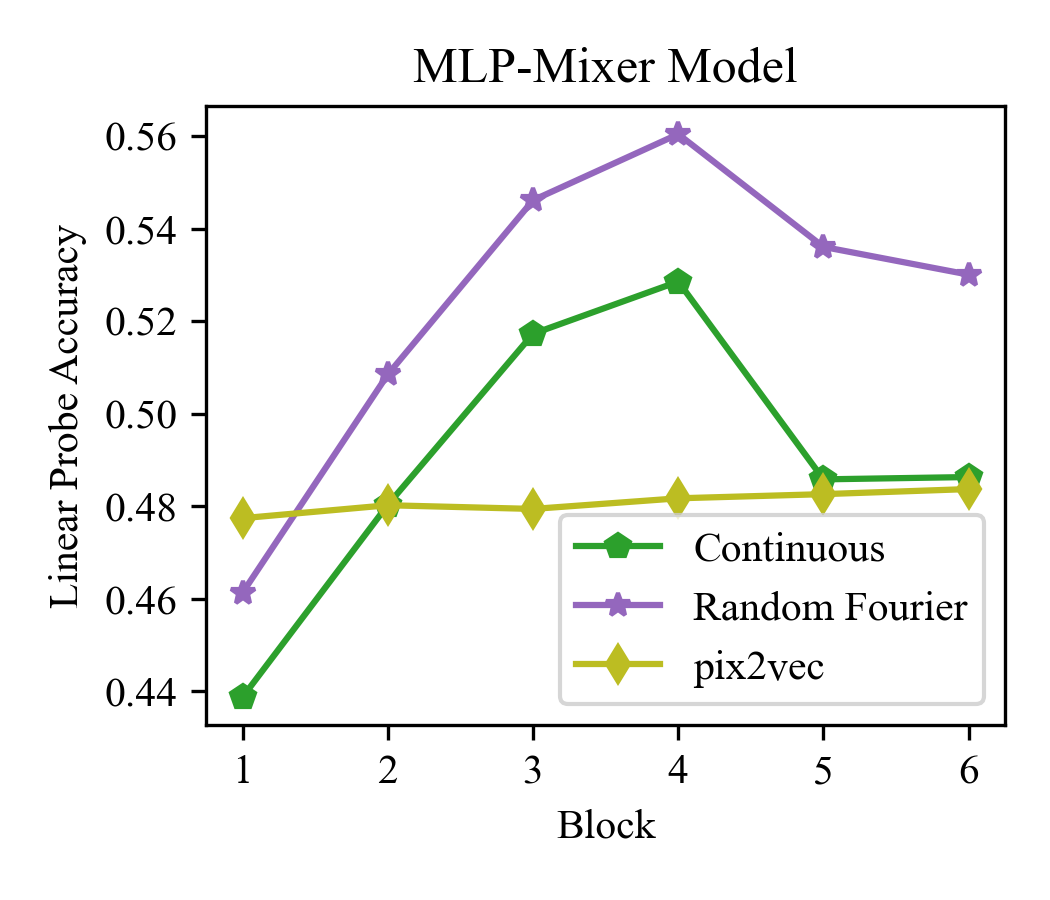} & \includegraphics[width=0.48\linewidth]{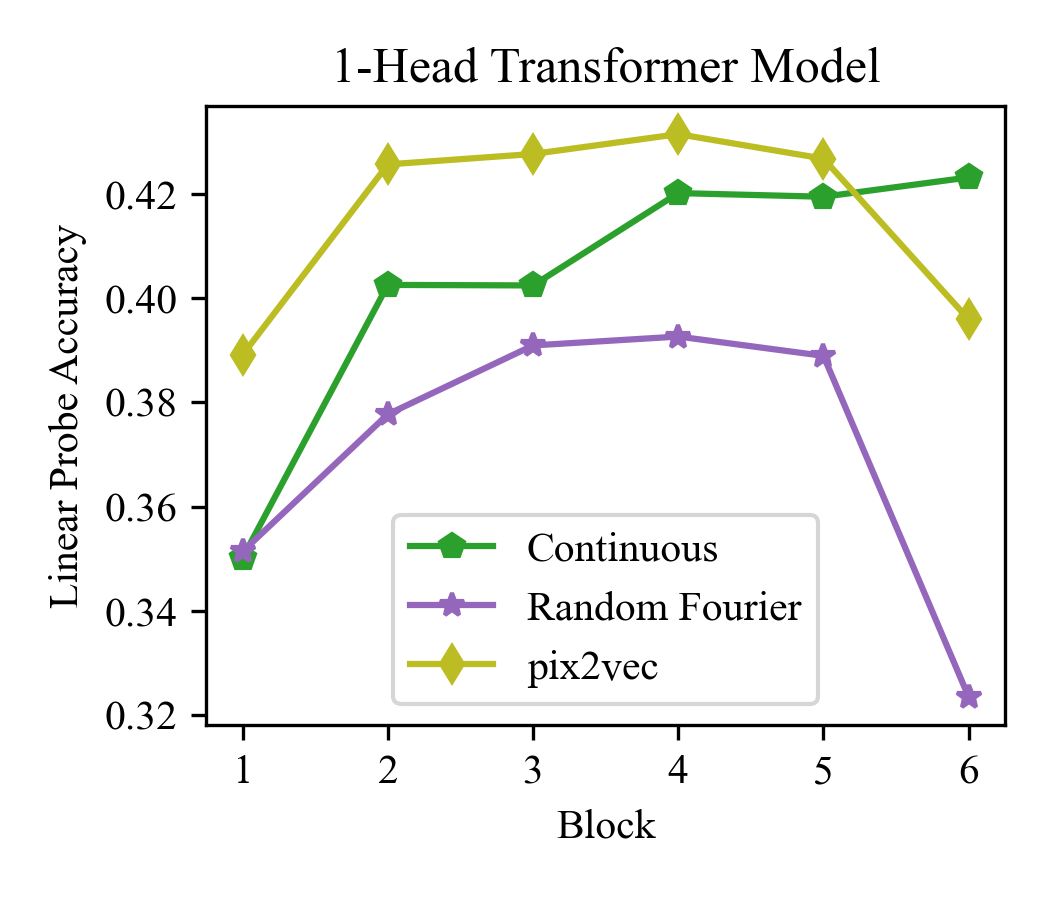}
    \end{tabular}
    \caption{(Left) We see that for the 6-block MLP-mixer, the random fourier network outperforms the continuous network at every layer while the pix2vec network performs equally bad at every layer. (Right) For the 6-block single-headed transformer, the pix2vec model actually outperforms the other two at all layers (except the final one)}
    \label{fig:depths}
\end{figure}
Looking at the right part of Figure \ref{fig:scatters}, we see that the best overall model in both terms of unsupervised performance and classification accuracy used the learned pix2vec embedding from the one-hot pixels. However, if we consider a fixed unsupervised loss, it appears the network using the continuous raw pixels would perform the best with the pix2vec network actually performing the worst. Furthermore, the classification performance of the random fourier encoded networks seems to lie in the space between that of the continuous and one-hot pix2vec encoded networks. Examining Figure \ref{fig:fourier_spectrum} we can see how lowering $\sigma$ would eventually lead to perfect distance preservation and raising $\sigma$ would converge to a random over-complete transformation of a one-hot embedding.

Having examined the overall trends of varying the input encoding method, we make further observations by examining the effect of varying the encoding method over a fixed architecture class. In the left part of Figure \ref{fig:depths}, we see that the random fourier encoded network actually outperforms the other two. Similarly, in the right part of Figure \ref{fig:depths}, we see that the one-hot pix2vec encoded network outperforms the others. These results come in contrast to the global trend of the continuous-encoded networks performing the best overall, suggesting that specific inductive biases (encoding and architectural) may pair better with each other than others.

\subsection{General Trends}
As stated earlier, one extremely noticeable trend from our experiments is the correlation between the performance of the unsupervised generative model and that of the down-stream classification task just as in image-GPT \cite{igpt}. This bodes well for the research community, as it implies breakthroughs discovered on small-scale isotropic unsupervised models may also transfer well to extremely large-scale models.

Another point of note is the abject failure of transformer models to compete with MLP-mixer or convolutional models as a whole. While transformer models are state-of-the-art for unsupervised visual learning with large-scale isotropic networks, it seems like this does not transfer well to small-scale problems. With our limited computing resources, it is difficult to determine if this is a data or architecture problem, though it is likely a combination of both.

Lastly, we identify the overall effect of inductive biases on our objective. Note that out of the encoding types, the raw continuous encoding performed the best on the classification task (with respect to unsupervised generative performance). Furthermore, out of the architecture classes, the convolutional networks outperformed both the MLP-mixer and transformer networks. Interestingly, this means that the two ``hardest'' inductive biases (locality in convolutions and perfect distance in continuous pixels) performed the best on our small-scale problem while both of these biases are removed in the extremely-large scale image-GPT \cite{igpt}. This supports the idea that small-scale networks do not have the capacity to re-learn these truths of nature from scratch and benefit greatly from having them explicitly baked into the model.
\section{Conclusions}
In this work, we study small-scale isotropic networks for unsupervised visual learning and analyze the impacts of the inductive biases that have been ablated out in extremely large-scale isotropic networks for the same task \cite{igpt}. While these massive networks can learn these truths from scratch over immense amounts of data, smaller networks seem to benefit greatly from having these inductive biases baked into the model. Specifically, we have shown that using a convolutional architecture with a continuous input outperforms using a transformer with one-hot input on small-scale networks, in contrast to large-scale networks like image-GPT.

Furthermore, our experimental results show that different types of input encodings yield better results with different types of isotropic network architectures, suggesting the possibility of as-of-yet untested inductive biases that may yield improved results when added to large-scale isotropic networks as well.

We have also shown that the general trend of a better unsupervised generative model yielding better results for down-stream classification translates well to small-scale isotropic networks regardless of model architecture or input encoding, suggesting that improvements made to the small-scale isotropic generative models may also lead to improvements for large-scale models as well.

Overall, we have provided insight as to the benefits of inductive biases in small-scale isotropic networks and illuminated ways for the typical AI researcher with modest compute to join in on current advancements and continue investigating these isotropic networks for pure unsupervised visual learning. Rather than throwing as much data and compute power as possible at an extremely large network, building a better understanding of the model's components, even at a small scale, will lead to quicker and greater advancements in the field.

\section{Acknowledgements}
This material is based upon work supported
by the National Science Foundation Graduate Research Fellowship under Grant No. DGE1745016.
\bibliography{neurips}

\end{document}